\documentclass{article}
\usepackage{amsmath}
\usepackage{amsfonts}
\usepackage{natbib}
\usepackage{booktabs}
\usepackage{multirow}
\usepackage{graphicx}  
\usepackage{here}
\usepackage{tikz}  
\usepackage{adjustbox}
\usepackage{multicol}

\usetikzlibrary{decorations.markings,fit,decorations.pathreplacing}
\usetikzlibrary{positioning,arrows.meta,calc,shapes.geometric,backgrounds,fit}
\usetikzlibrary{calc}
\title{Channel-Wise MLPs Improve the Generalization of Recurrent Convolutional Networks}
\author{Nathan Breslow}
\date{\today}

\begin{document}
\maketitle

\section {Abstract}

We investigate the impact of channel-wise mixing via multi-layer perceptrons (MLPs) on the generalization capabilities of recurrent convolutional networks. Specifically, we compare two architectures: DARC (Depth Aware Recurrent Convolution), which employs a simple recurrent convolutional structure, and DAMP (Depth Aware Multi-layer Perceptron), which extends DARC with a gated MLP for channel mixing. Using the Re-ARC benchmark, we find that DAMP significantly outperforms DARC in both in-distribution and out-of-distribution generalization under exact-match grading criteria. These results suggest that explicit channel mixing through MLPs enables recurrent convolutional networks to learn more robust and generalizable computational patterns. Our findings have implications for neural program synthesis and highlight the potential of DAMP as a target architecture for hypernetwork approaches.

\section {Introduction}
The Abstraction and Reasoning Corpus (ARC) challenge \citep{chollet2019measure} has sparked renewed interest in neural architectures capable of program-like computation and symbolic manipulation. Unlike traditional deep learning benchmarks that rely on interpolation within massive training distributions, ARC tasks require discovering and applying abstract rules from just a few examples. This has led researchers to explore neural program synthesis approaches \citep{victorvikram2020arcicecuber, greenblatt2024getting50, li2025combining}, where networks learn to generate executable programs or perform systematic transformations. A key insight from this line of work is that successful architectures must handle variable-sized inputs while maintaining strong generalization capabilities—properties that extend well beyond ARC to many vision and reasoning tasks.

Among architectures capable of such generalization, recent work has shown that recurrent convolutional networks show promise. \citet{schwarzschild2021can} demonstrated that such architectures could successfully scale from 9×9 to 13×13 mazes and achieve performance gains on chess puzzles through deeper test-time iteration. Similarly, adaptive depth mechanisms in Transformers have shown strong generalization abilities on tasks like binary summation and multiplication \citep{fan2024looped}.

Building on these insights, we investigate whether augmenting depth-aware recurrent convolutions with channel-wise mixing can improve generalization performance. Specifically, we compare two architectures: DARC (Depth Aware Recurrent Convolution), which uses a simple recurrent convolutional structure, and DAMP (Depth Aware Multi-layer Perceptron), which extends DARC with a gated MLP for channel mixing.

Our key finding is that despite the minimal architectural difference - DAMP simply adds a channel-wise gated MLP after the convolution - this modification leads to substantial improvements in both in-distribution and out-of-distribution generalization. Using the Re-ARC benchmark as our testbed, we show that DAMP achieves a median accuracy of 92.19\% on in-distribution tasks compared to DARC's 78.75\%, and more remarkably, 14.58\% on out-of-distribution tasks compared to DARC's 2.34\%, in a parameter-matched setting.

This work makes two primary contributions:
\begin{enumerate}
\item We demonstrate that channel-wise MLPs provide a surprisingly effective mechanism for improving the generalization capabilities of recurrent convolutional networks, particularly for out-of-distribution inputs.
\item We validate that both architectures can achieve strong performance across 185 diverse tasks using a single fixed network configuration, suggesting their potential as target architectures for future hypernetwork approaches.
\end{enumerate}

The significance of these results extends beyond the specific benchmark: they suggest that explicit channel mixing through MLPs enables recurrent convolutional networks to learn more robust and generalizable computational patterns, even when the base architecture already performs implicit channel mixing through convolutions.

\section {DARC and DAMP}

DARC (Depth Aware Recurrent Convolution) consists of a simple linear embedding layer followed by a looped convolution that iteratively refines the input grid by adding to a residual stream. This recurrent block is repeated $N$ times, where $N$ is determined from the size of the input grid - typically $N = 2 * max(H_i, W_i)$, where $H_i$ and $W_i$ are the height and width of the input grid. At the last iteration, the residual stream is passed to the transpose of the embedding layer's weight matrix to produce the output grid. The architecture is illustrated below:

\definecolor{embedcolor}{RGB}{255,179,102}     
\definecolor{convcolor}{RGB}{135,206,235}      
\definecolor{normcolor}{RGB}{144,238,144}      
\definecolor{actcolor}{RGB}{221,204,255}       
\definecolor{mlpcolor}{RGB}{255,128,149}       
\definecolor{linearcolor}{RGB}{255,204,153}    
\definecolor{outcolor}{RGB}{177,156,217}       
\definecolor{inputcolor}{RGB}{255,239,153}     
\definecolor{outputcolor}{RGB}{255,204,229}    

\tikzstyle{block} = [rectangle, draw, thick, black, minimum width=3.2cm, minimum height=1cm, align=center, rounded corners=8pt]
\tikzstyle{input} = [block, fill=inputcolor]
\tikzstyle{output} = [block, fill=outputcolor]
\tikzstyle{embed} = [block, fill=embedcolor]
\tikzstyle{conv} = [block, fill=convcolor]
\tikzstyle{act} = [block, fill=actcolor]
\tikzstyle{linear} = [block, fill=linearcolor]
\tikzstyle{norm} = [block, fill=normcolor]
\tikzstyle{mlp} = [block, fill=mlpcolor]
\tikzstyle{out} = [block, fill=outcolor]
\tikzstyle{arrow} = [-{Stealth[scale=1.3]}, thick, black]
\tikzstyle{residual} = [arrow]
\tikzset{arrow/.style={-{Stealth[scale=1.3]}, thick, black}}

\adjustbox{scale=0.75}{

\begin{tikzpicture}[node distance=0.6cm and 0.8cm][H]
    \node[font=\huge\bfseries] at (0, 12) {DARC Model};
    
    \node[input] (input) at (0, 0) {Grid Inputs \quad \\{\small $(B, H, W)$}};
    \draw[arrow] (0, 0.5) -- (0, 1);
    
    \node[embed] (embed) at (0, 1.8) {Input\\Embedding};
    
    
    \node[draw, thick, rounded corners=15pt, dashed, 
            minimum width=5.0cm, minimum height=5.25cm, 
            fill=gray!5,
            postaction={decorate,
            decoration={markings,
                mark=at position 0.15 with {\arrowreversed{Stealth[scale=1.3]}},
                mark=at position 0.65 with {\arrowreversed{Stealth[scale=1.3]}},
            }
            }] (middlebox) at (0, 5.5) {};

    \node at (-4.5, 6.8) {\Large $\left[2 \cdot \operatorname{max}(H,W)\right]\times$};
    
    \coordinate (blockstart) at (0, 3.3);
    
    \node[norm] (addnorm1) at (0, 4) {LayerNorm};
    
    \node[conv] (conv) at (0, 5.5) {Conv2d\\{\small $3\times3$}};
    
    \node[act] (gelu) at (0, 7) {GeLU};
    
    \coordinate (blockend) at (0, 8.125);
    
    \draw[arrow] (embed) -- (addnorm1);
    \draw[arrow] (addnorm1) -- (conv);
    \draw[arrow] (conv) -- (gelu);
    \draw[thick] (gelu) -- (blockend);
    
    \node[font=\fontsize{22}{22}\bfseries] at (blockend) {$\oplus$};   
    
    \draw[arrow] (blockend) -- ++(0, 1.5);
    
    \node[embed] (outembed) at (0, 9.3) {Output Head};
    
    \node[output] (output) at (0, 10.8) {Output Logits \quad \\{\small $(B, H, W, C)$}};

    \draw[arrow] (outembed) -- (output);

    \draw[<->, very thick, blue!60] 
        (embed.east) .. controls (4, 1.8) and (4, 9.3) .. 
        node[midway, right, blue!60] {$W^T$} (outembed.east);
\end{tikzpicture}
}

The DARC architecture additionally has several benefits: it can accept variable input grid sizes, it's easily interpretable via the residual stream, and (as will be shown) it retains expressive functionality at extremely small scales.

DAMP (Depth Aware Multi-layer Perceptron) is a variant of DARC that follows up the depth-aware convolution with a multi-layer perceptron (MLP) that processes the output of the recurrent convolution channel-wise. The MLP follows the GeLU gating established in \citep{Shazeer2020}:

$\mathrm{FFN}_{\mathrm{GEGLU}}(x, W, V, W_2) = (\mathrm{GELU}(xW) \otimes xV) W_2$

This allows rich expressivity in how the model can mix channel information. Without the MLP, the naive DARC model can only take a simple linear combination of the channels in the output grid and threshold it with an activation function. While this can theoretically approximate any function the MLP could learn (if the network was sufficiently wide), the MLP may potentially provide a more efficient and effective way to learn complex channel interactions.

The full DAMP architecture is illustrated below:

\medskip
\adjustbox{scale=0.75}{

\begin{tikzpicture}[node distance=0.6cm and 0.8cm][H]

    \node[font=\huge\bfseries] at (0, 17) {DAMP Model};
    \node[input] (input) at (0, 0) {Grid Inputs \quad \\{\small $(B, H, W)$}};
    \draw[arrow] (0, 0.5) -- (0, 1);
    
    \node[embed] (embed) at (0, 1.8) {Input\\Embedding \\{\small $Wx$}};
    

    \node at (-4.5, 8.0625) {\Large $\left[2 \cdot \operatorname{max}(H,W)\right]\times$};
    \coordinate (blockstart) at (0, 3.3);
    
    \node[norm] (addnorm1) at (0, 4) {LayerNorm};
    
    \node[conv] (conv) at (0, 5.5) {Conv2d\\{\small $3\times3$}};
    
    \node[act] (gelu) at (0, 7) {GeLU};
    
    \coordinate (blockend) at (0, 8.125);
    
    \draw[arrow] (embed) -- (addnorm1);
    \draw[arrow] (addnorm1) -- (conv);
    \draw[arrow] (conv) -- (gelu);
    
    \node[font=\fontsize{22}{22}\bfseries] (resid1) at (blockend) {$\oplus$};   
    \node[norm] (addnorm2) at (0, 9.5) {LayerNorm};
    \draw[arrow] (blockend) -- (addnorm2);
    \node[mlp] (mlp) at (0, 11) {Gated MLP};
    \draw [arrow] (addnorm2) -- (mlp);
    \node[input] (mlpinput) at (6, 9.5) {MLP Input \quad \\{\small $(\dots, C)$}};
    \node[linear] (up_proj) at (4, 11) {Linear Layer \quad \\{\small $(\dots, C) \rightarrow (\dots, 4C)$}};
    \node[linear] (gate_proj) at (8, 11) {Linear Layer \quad \\{\small $(\dots, C) \rightarrow (\dots, 4C)$}};
    \draw[arrow] (mlpinput) -- (up_proj);
    \draw[arrow] (mlpinput) -- (gate_proj);
    \node[font=\fontsize{22}{22}\bfseries] (gate_times) at (6, 12) {$\odot$};   
    \draw[arrow] (up_proj) |- (gate_times);
    \draw[arrow] (gate_proj) |- (gate_times);
    \node[linear] (down_proj) at (6, 13.5) {Linear Layer \quad \\{\small $(\dots, 4C) \rightarrow (\dots, C)$}};
    \draw[arrow] (gate_times) -- (down_proj);
    \node[output] (output) at (6, 15) {MLP Output \quad \\{\small $(\dots, C)$}};
    \draw[arrow] (down_proj) -- (output);
    \node[fit=(mlpinput)(down_proj)(up_proj)(gate_proj)(gate_times)(output), inner sep=0.3cm] (mlpfit) {};
    \draw[decorate,decoration={brace,amplitude=10pt,mirror,aspect=0.675}] 
        (mlpfit.north west) -- (mlpfit.south west) 
        node[midway,align=center] {};
    \draw[arrow] (gelu) -- ($(resid1) + (0, -0.2)$);

    \node[font=\fontsize{22}{22}\bfseries] (resid2) at (0, 12.25) {$\oplus$}; 
    \draw[arrow] ($(embed) + (0, 1)$) -- +(-2, 0) |- (resid1);
    \draw[arrow] ($(resid1) + (0, 0.375)$) -- +(-2, 0) |- (resid2);
    \draw[arrow] (mlp) -- (resid2);

    \node[fit=(resid1)(addnorm1)(resid2), inner sep=0.3cm] (residfit) {};
    \draw[decorate,decoration={brace,amplitude=10pt,mirror,aspect=0.5}] 
        ([xshift=-5pt]residfit.north west) -- ([xshift=-10pt]residfit.south west) 
        node[midway,align=center] {};

    
    \node[embed] (outembed) at (0, 13.75) {Output Head \quad \\{\small $W^Tx$}};
    
    \node[output] (output) at (0, 15.25) {Output Logits \quad \\{\small $(B, H, W, C)$}};

    \draw[arrow] (resid2) -- (outembed);
    \draw[arrow] (outembed) -- (output);

\end{tikzpicture}
}
\section {Methodology}

We employ the Re-ARC dataset \citep{hodel2024addressingabstractionreasoningcorpus} as our primary benchmark for evaluating the generalization capabilities of DARC and DAMP. Re-ARC is a procedurally generated dataset that provides a generator function for each of the 400 ARC training tasks. Each generator has a configurable difficulty range, which allows us to sample tasks of lower difficulty for training and higher difficulty for out-of-distribution (OOD) evaluation. This lets us systematically assess how well models generalize to more complex tasks than those seen during training across a wide variety of transformations.

Each generator takes the form:

$f(l, u) \rightarrow (G_i, G_o), \quad l, u \in [0, 1], \quad G_i \in [C]^{H_i \times W_i}, \quad G_o \in [C]^{H_o \times W_o}$

where $l$ and $u$ are the lower and upper bounds of the difficulty range, $G_i$ is the input grid, $G_o$ is the output grid, and $C$ is the number of channels in the grid.

185 Re-ARC tasks are selected from the Re-ARC dataset of 400 tasks. These tasks satisfy two principal requirements - they preserve the size of the grid they act on ($H_i = H_o, W_i = W_o$), and they generate at more than 200 tasks/second on a single M3 Max CPU core.
For each task, we create two new model instances:
\begin{itemize}
\item DARC architecture: $\theta_{\text{darc}}$ with intermediate size 98 (87.91K parameters)
\item DAMP architecture: $\theta_{\text{damp}}$ with intermediate size 64 (87.55K parameters)
\end{itemize}

Both models are randomly initialized from scratch for every task with a different random seed.

Training setup:
\begin{itemize}
\item Train for 2048 batches of 16 examples each
\item Each batch $B$ is generated as: $B \sim f(0.3, 0.5)_i$ (examples sampled with difficulty between 0.3 and 0.5)
\item Total examples trained on: $2048 \times 16 = 32,768$
\end{itemize}

Optimization:
\begin{itemize}
\item Muon optimizer with learning rate $\eta = 0.01$
\item 16 warmup steps
\item Cosine decay to $\eta_{\text{final}} = 10^{-7}$
\end{itemize}

Loss function:
\begin{itemize}
\item Cross entropy loss between model output $\theta(G_i)$ and target $G_o$:
$$\mathcal{L} = \text{CrossEntropy}(\theta(G_i), G_o)$$
\end{itemize}
\pagebreak
\section {Results}
For each learned task we sample difficulty values over the ID or OOD range in increments of 0.05. At every difficulty value we generate 64 grid instances and score each by exact match. We average these 64 scores to obtain a difficulty-level accuracy, then average across all sampled difficulty values to obtain the task’s accuracy. Overall test accuracy is the mean of these per-task accuracies across the 185 benchmark tasks.

Test Accuracy (ID) corresponds to a Re-ARC difficulty range of $0.3-0.5$ and Test Accuracy (OOD) corresponds to a Re-ARC difficulty range of $0.55-0.8$. The difference between the training distribution and the testing distribution can be quite dramatic for some tasks - see the case study of 4938f0c2 below:

\begin{figure}[H]
    \centering
    \includegraphics[width=0.8\textwidth]{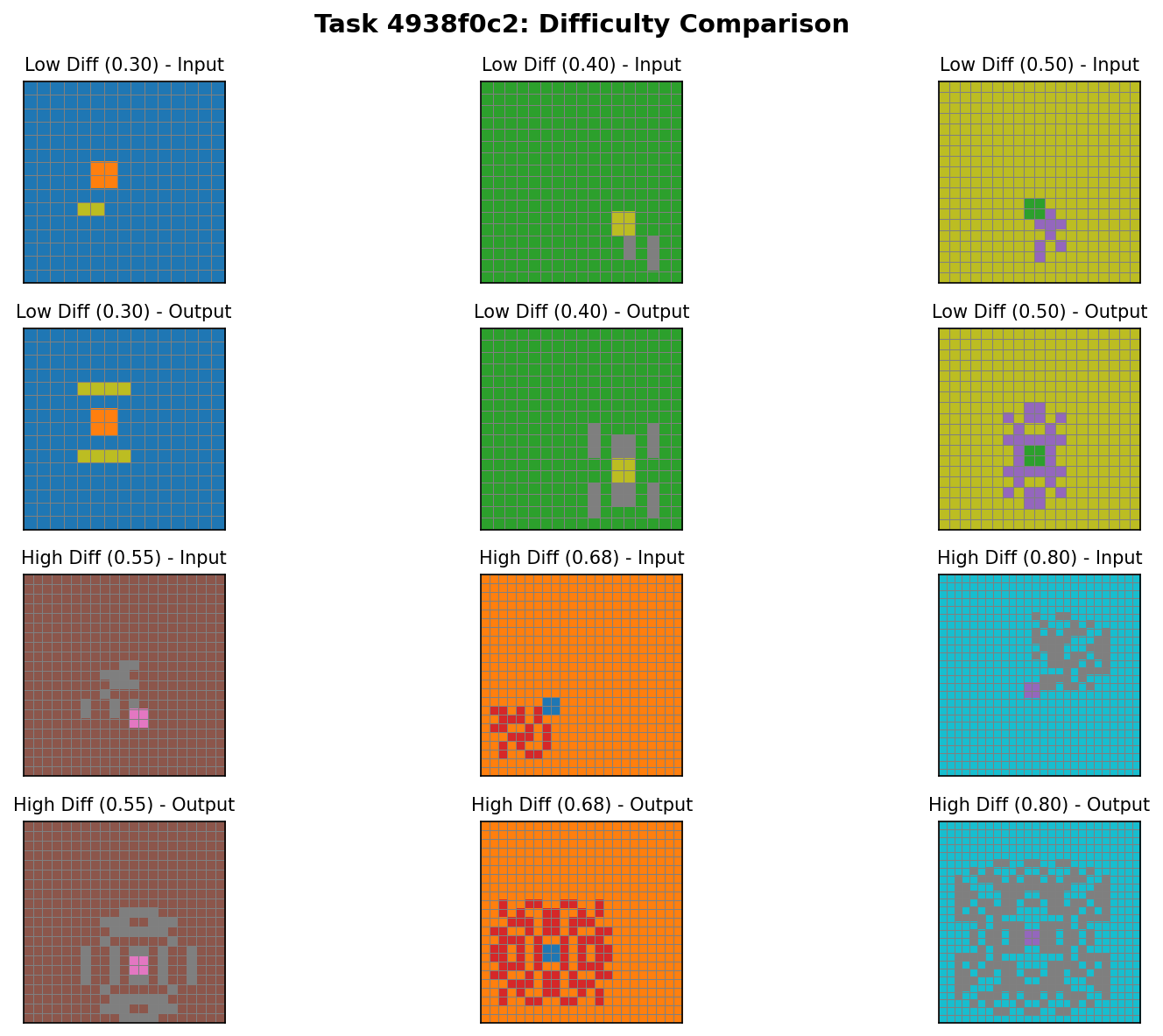}
    \caption{Difficulty across 0.3-0.8 range for the for task 4938f0c2, which performs a nontrivial transformation on the input grid.}
    \label{fig:4938f0c2_difficulty_vs_accuracy}
\end{figure}
Note that all grids used for computing both Test (ID) and Test (OOD) accuracies were not seen by the model during training, as they are sampled from the generator function after training is complete. The results of DARC vs.\ DAMP on the 185-task Re-ARC benchmark are summarized in Table \ref{tab:darc_damp_comprehensive} below.

\begin{table}[H]
\centering
\caption{DARC vs.\ DAMP performance on the 185-task Re-ARC benchmark.  
Bold numbers mark the better value within each ID or OOD column pair. All 95\% CIs are percentile-bootstrap
with $B=10{,}000$.}
\label{tab:darc_damp_comprehensive}
\begin{tabular}{lcccc}
\toprule
\multirow{2}{*}{Metric} & \multicolumn{2}{c}{Test Accuracy (ID)} & \multicolumn{2}{c}{Test Accuracy (OOD)} \\
\cmidrule(lr){2-3} \cmidrule(lr){4-5}
 & DARC & DAMP & DARC & DAMP \\
\midrule
\multicolumn{5}{l}{\textbf{Summary Statistics}} \\
Mean                 & 54.77\% & \textbf{60.21\%} & 31.08\% & \textbf{38.98\%} \\
Median               & 78.75\% & \textbf{92.19\%} & 2.34\% & \textbf{14.58\%} \\
Mean Diff (DAMP$-$DARC) & \multicolumn{2}{c}{+5.43\%} & \multicolumn{2}{c}{+7.90\%} \\
Diff 95\% CI         & \multicolumn{2}{c}{[2.71\%, 8.18\%]} & \multicolumn{2}{c}{[5.56\%, 10.32\%]} \\
\midrule
\multicolumn{5}{l}{\textbf{Win/Loss/Tie Analysis}} \\
DAMP $>$ DARC        & \multicolumn{2}{c}{80 (43.2\%)} & \multicolumn{2}{c}{89 (48.1\%)} \\
DAMP $=$ DARC $=$ 1  & \multicolumn{2}{c}{39 (21.1\%)} & \multicolumn{2}{c}{7 (3.8\%)}  \\
DAMP $=$ DARC $=$ 0  & \multicolumn{2}{c}{42 (22.7\%)} & \multicolumn{2}{c}{71 (38.4\%)} \\
DAMP $=$ DARC (other)& \multicolumn{2}{c}{5 (2.7\%)}  & \multicolumn{2}{c}{1 (0.5\%)}  \\
DAMP $<$ DARC        & \multicolumn{2}{c}{19 (10.3\%)} & \multicolumn{2}{c}{17 (9.2\%)} \\
\midrule
\multicolumn{5}{l}{\textbf{Statistical Tests}} \\
Paired $t$-test      & \multicolumn{2}{c}{$p < 0.001$} & \multicolumn{2}{c}{$p < 0.001$} \\
Wilcoxon test        & \multicolumn{2}{c}{$p < 0.001$} & \multicolumn{2}{c}{$p < 0.001$} \\
Cohen’s $d$          & \multicolumn{2}{c}{0.281}       & \multicolumn{2}{c}{0.481}       \\
Cliff's $\delta$     & \multicolumn{2}{c}{0.0771}       & \multicolumn{2}{c}{0.0890}       \\
\midrule
\multicolumn{5}{l}{\textit{Total tasks: 185}} \\
\bottomrule
\end{tabular}
\end{table}
The table above reveals that DAMP consistently outperforms DARC both in-distribution (ID) and out-of-distribution (OOD) across all metrics. The mean accuracy difference is statistically significant (p $<$ 0.001), with DAMP achieving a higher mean accuracy in both ID and OOD settings. The win/loss/tie analysis shows that DAMP wins over DARC in 43.2\% of ID tasks and 48.1\% of OOD tasks.

The most striking performance differences are observed in median accuracy and out-of-distribution (OOD) performance. The median accuracy for DARC is 78.75\% in ID tasks, while DAMP achieves a significantly higher median of 92.19\%. In OOD tasks, the contrast is far more profound - DARC's median accuracy is only 2.34\%, while DAMP achieves 14.58\% - a standout difference. 

Both methods, however, are still limited - they both fail completely on 22.7\% of ID tasks and 38.4\% of OOD tasks. Mean accuracy on OOD tasks is still below 40\% for both methods.

Regardless, the exceptional performance of DAMP OOD (in comparison to DARC) is further supported by the difference in Cohen's d values, which indicate a small effect size for ID tasks (0.281) and a medium effect size for OOD tasks (0.481). However, the Cliff's $\delta$ values suggest that the practical significance of these differences is relatively negligible, with values of 0.0771 for ID and 0.0890 for OOD tasks. This is incongruent with the rest of the analysis, but the reasons for this discrepancy are elucidated when considering the underlying distribution. The histogram below shows that most performance gains and failures come from either completely failing to solve a task or solving it perfectly. The distribution of accuracies is bimodal, with peaks at 0\% and 100\% accuracy.

\begin{figure}[H]
    \centering
    \includegraphics[width=0.8\textwidth]{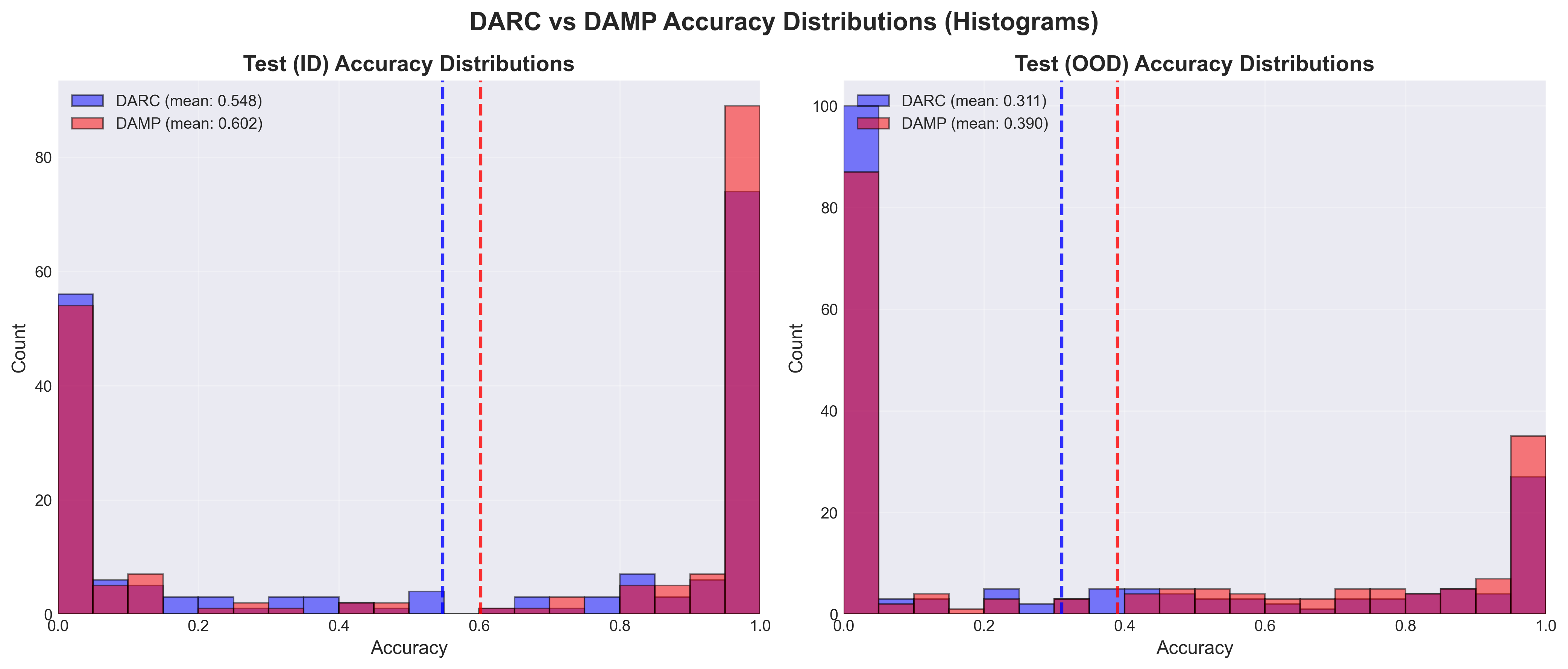}
    \caption{Histogram of DARC and DAMP accuracies on 185 Re-ARC tasks. The bimodal distribution indicates that most tasks are either solved perfectly or not at all.}
    \label{fig:darc_damp_histogram}
    \end{figure}

This bimodal pattern explains the low Cliff's $\delta$ values. Since Cliff's $\delta$ measures the probability that a randomly selected value from one distribution exceeds a value from another distribution, it remains small when only a few tasks shift from the "unsolved" (0\%) to "solved" (100\%) category. 

The reason this unique pattern emerges is the exact-match grading criteria - there is no room for error. A single pixel mismatch results in a zero score. This is important when contextualizing the results - the mean accuracy improvement from 31.08\% to 38.98\% for OOD tasks corresponds to an expected 7.90\% increase in the number of grids solved entirely correctly for an \textit{arbitrary} ARC task. Furthermore, this improvement holds up under scrutiny - the 95\% confidence intervals for the mean difference between DAMP and DARC exclude 0 for both ID and OOD tasks, indicating that the observed performance gains are statistically significant. These intervals are computed via bootstrapping, which is appropriate given the non-normal distribution of accuracies.

What makes this fundamentally surprising is the incredible similarity of DARC and DAMP - DAMP simply adds a singular gated MLP applied channel-wise. 3x3 convolutions \textit{already} mix channels nonlinearly, so the addition of a gated MLP shouldn't have a tremendous impact on DARC's expressiveness. However, the results show that DAMP is able to learn more complex mappings and generalize better to unseen tasks, especially in the OOD setting. Simply allocating additional compute to the processing of channels between convolutions allowed a surprisingly large performance boost in parameter-matched settings.

\section {Discussion}

The results presented here demonstrate that augmenting recurrent convolutional architectures with channel-wise MLPs can significantly enhance their generalization capabilities, particularly in out-of-distribution scenarios. The DAMP architecture, which incorporates a gated MLP after the depth-aware convolution, consistently outperforms the simpler DARC model across a diverse set of tasks from the Re-ARC benchmark. The exact reasons as to why DAMP performs so much better than DARC are still unclear. It is possible that the MLP provides a more efficient way to learn complex channel interactions, or that it helps the model learn more robust features that generalize better to unseen tasks. Further analysis is needed to understand the underlying mechanisms driving this performance difference.

These findings have important implications for making progress on ARC and similar challenges that require strong generalization from limited data. Specifically, the proposed architectures offer a novel avenue into neural program synthesis.

Specifically, proposed neural program synthesis \citep{li2025combining, greenblatt2024getting50} approaches typically involve a network generating Python code that is then executed to produce the output grid. The main issue with this approach is twofold - first, not all ARC tasks have easily discrete solutions in Python. Some rely on intuition and pattern recognition that is difficult to express in code. Second, Python programs are not differentiable, so they cannot benefit directly from test-time training on provided examples - a technique demonstrated to be extremely effective on ARC tasks \citep{akyürek2025surprisingeffectivenesstesttimetraining}.

An alternative method of neural program synthesis is the hypernetwork - a network that, given few-shot examples, generates the weights of a target network that can solve the task. This target network can then be tuned on the provided examples, allowing for test-time training. The challenge with this approach is finding a target architecture that is both expressive enough to solve a wide variety of tasks and small enough to be generated by a hypernetwork. The results here suggest that DAMP is a promising candidate for such a target architecture, as it can achieve strong performance across 185 diverse tasks using a single fixed network configuration. It could be extended to support transformations that don't preserve grid size through simple padding, and its recurrent nature allows it to handle variable-sized inputs. Hypernetworks have already been shown to be effective at the few-shot generation of weights for classification tasks in NLP \citep{volk2023examplebasedhypernetworksoutofdistributiongeneralization}, so it is plausible that they could generate weights for DAMP to solve ARC tasks given few-shot examples. This direction presents an exciting avenue for future research.

Finally, we must clarify that our demonstration does not constitute a general solution to ARC-AGI. A true solution would need to show few-shot generalization to completely new tasks, rather than what we've shown here: strong out-of-distribution performance on a single task after extensive training on simpler versions of that same task. Regardless, this generalization provides important evidence that DAMP's inductive biases, relative to naive recurrent convolutional networks, are well-aligned with the types of reasoning patterns needed for ARC.

\section {Conclusion}

In conclusion, this work highlights the significant impact of channel-wise MLPs on the generalization capabilities of recurrent convolutional networks. The DAMP architecture, with its simple yet effective modification to DARC, demonstrates that even small architectural changes can lead to substantial performance improvements, particularly in out-of-distribution settings. These findings open up new possibilities for neural program synthesis and suggest that DAMP could serve as a robust target architecture for hypernetwork approaches. Future work will focus on further analyzing the mechanisms behind DAMP's success and exploring its integration into neural program synthesis frameworks.

\section*{Acknowledgements}

Following the example of \citet{franzen2025productexpertsllmsboosting}, we acknowledge LLMs were used during the ideation, drafting, writing, and editing of this paper and assume full responsibility for all content.

\newpage

\bibliographystyle{plainnat}
\bibliography{main}

\section*{Appendix}

\subsection {Optimizer Ablation}

The Muon optimizer was chosen for this study due to preliminary evaluations showing it strongly outperformed baseline AdamW settings. However, the advantage DAMP has over DARC remains even when using a naive AdamW configuration. The table below shows the results of training DARC and DAMP with AdamW instead of Muon, confirming that the performance difference between the two architectures is not due to the choice of optimizer. AdamW was trained with a learning rate of 0.001, a $\beta_1$ of 0.9, a $\beta_2$ of 0.999, and a weight decay of 0.01.

\begin{table}[H]
\centering
\caption{Performance comparison of DARC and DAMP using Muon vs AdamW optimizers on the 185-task Re-ARC benchmark. All models were trained with identical hyperparameters except for the optimizer choice. Due to the non-normal distribution of accuracies, we report medians alongside means and use Wilcoxon signed-rank tests.}
\label{tab:optimizer_ablation}
\begin{tabular}{lcccc}
\toprule
\multirow{2}{*}{Model-Optimizer} & \multicolumn{2}{c}{Mean Accuracy} & \multicolumn{2}{c}{Median Accuracy} \\
\cmidrule(lr){2-3} \cmidrule(lr){4-5}
 & Test (ID) & Test (OOD) & Test (ID) & Test (OOD) \\
\midrule
DARC-Muon & \textbf{54.77\%} & \textbf{31.08\%} & \textbf{78.75\%} & \textbf{2.34\%} \\
DAMP-Muon & \textbf{60.21\%} & \textbf{38.98\%} & \textbf{92.19\%} & \textbf{14.58\%} \\
DARC-AdamW & 15.27\% & 5.28\% & 0.00\% & 0.00\% \\
DAMP-AdamW & 21.26\% & 9.04\% & 0.00\% & 0.00\% \\
\midrule
\multicolumn{5}{l}{\textbf{Relative Improvement from Muon (based on means)}} \\
DARC & \multicolumn{2}{c}{ID: +258.6\%} & \multicolumn{2}{c}{OOD: +489.0\%} \\
DAMP & \multicolumn{2}{c}{ID: +183.2\%} & \multicolumn{2}{c}{OOD: +331.3\%} \\
\midrule
\multicolumn{5}{l}{\textbf{Statistical Significance (Wilcoxon signed-rank tests, Test-ID)}} \\
Muon $>$ AdamW (DARC) & \multicolumn{2}{c}{$p < 0.001$} & \multicolumn{2}{c}{Cliff's $\delta$ = 0.521 (large)} \\
Muon $>$ AdamW (DAMP) & \multicolumn{2}{c}{$p < 0.001$} & \multicolumn{2}{c}{Cliff's $\delta$ = 0.486 (large)} \\
DAMP $>$ DARC (Muon) & \multicolumn{2}{c}{$p < 0.001$} & \multicolumn{2}{c}{Cliff's $\delta$ = 0.077 (negligible)} \\
DAMP $>$ DARC (AdamW) & \multicolumn{2}{c}{$p < 0.001$} & \multicolumn{2}{c}{Cliff's $\delta$ = 0.087 (negligible)} \\
\bottomrule
\end{tabular}
\end{table}

As the table above shows, even with the AdamW optimizer, DAMP still outperforms DARC by a significant margin - with a mean of $9.04\%$ vs.\ $5.28\%$ on OOD tasks when trained with AdamW.  The Cliff's $\delta$ values - previously established to be biased downward due to exact-match grading criteria - still indicate large effect sizes for the performance difference between Muon and AdamW optimizers.

This validates both the choice of Muon as the optimizer for this study and the robustness of the performance difference between DARC and DAMP. Note that compute limitations prevented a full hyperparameter sweep for AdamW, so this ablation simply serves to show why we chose Muon as the optimizer for this study - it provided better performance than AdamW on both DARC and DAMP and therefore allowed a more nuanced evaluation. We do not have the evidence to claim that, in general, Muon is a better optimizer than AdamW for ARC tasks.

\subsection {Chosen Tasks}

The following 185 tasks were selected from the Re-ARC dataset for this study. These tasks were chosen based on their ability to preserve grid size and generate at more than 200 tasks/second on a single M3 Max CPU core. The full list of tasks is as follows:

\begin{multicols}{4}  
\footnotesize  
\begin{enumerate}
\item d4a91cb9
\item e9614598
\item 8403a5d5
\item a2fd1cf0
\item b8cdaf2b
\item 67a423a3
\item 25d8a9c8
\item d90796e8
\item f76d97a5
\item 6e02f1e3
\item 2281f1f4
\item dc433765
\item 3bd67248
\item 6f8cd79b
\item 928ad970
\item e48d4e1a
\item c1d99e64
\item a699fb00
\item ea786f4a
\item 834ec97d
\item d406998b
\item d037b0a7
\item 2204b7a8
\item 952a094c
\item 8f2ea7aa
\item 1e0a9b12
\item 4612dd53
\item 5582e5ca
\item 3ac3eb23
\item ba97ae07
\item 7ddcd7ec
\item 54d82841
\item 4c5c2cf0
\item 4938f0c2
\item a5f85a15
\item 8eb1be9a
\item 1f876c06
\item f15e1fac
\item dc1df850
\item aba27056
\item 29623171
\item 1caeab9d
\item 4258a5f9
\item 08ed6ac7
\item 99fa7670
\item dbc1a6ce
\item a85d4709
\item 3c9b0459
\item d9f24cd1
\item 6150a2bd
\item ec883f72
\item e9afcf9a
\item ba26e723
\item 67a3c6ac
\item 32597951
\item 68b16354
\item d4f3cd78
\item 9dfd6313
\item c444b776
\item 7447852a
\item 3906de3d
\item c9f8e694
\item 97999447
\item bdad9b1f
\item aedd82e4
\item ded97339
\item 673ef223
\item ce22a75a
\item e21d9049
\item 1e32b0e9
\item e8dc4411
\item 42a50994
\item 40853293
\item 3618c87e
\item 2bee17df
\item a9f96cdd
\item 913fb3ed
\item d23f8c26
\item 56ff96f3
\item a8d7556c
\item 28e73c20
\item 0ca9ddb6
\item fcc82909
\item ae3edfdc
\item a48eeaf7
\item 85c4e7cd
\item 321b1fc6
\item 4093f84a
\item e76a88a6
\item 93b581b8
\item 045e512c
\item 50cb2852
\item 82819916
\item 0962bcdd
\item f1cefba8
\item ce9e57f2
\item d89b689b
\item 67385a82
\item b1948b0a
\item d364b489
\item 3bdb4ada
\item 88a10436
\item cbded52d
\item 694f12f3
\item 9565186b
\item 54d9e175
\item 4347f46a
\item b60334d2
\item 95990924
\item e8593010
\item f35d900a
\item 623ea044
\item bda2d7a6
\item 1a07d186
\item 6d0160f0
\item 31aa019c
\item db3e9e38
\item 6cdd2623
\item b27ca6d3
\item d5d6de2d
\item 05f2a901
\item 178fcbfb
\item 5168d44c
\item 23581191
\item 22eb0ac0
\item 3aa6fb7a
\item b782dc8a
\item e5062a87
\item 253bf280
\item 05269061
\item d511f180
\item bb43febb
\item c8f0f002
\item ecdecbb3
\item 09629e4f
\item a78176bb
\item 7b6016b9
\item d6ad076f
\item b6afb2da
\item 41e4d17e
\item 868de0fa
\item 91714a58
\item 6cf79266
\item 2dd70a9a
\item 2c608aff
\item b8825c91
\item 9edfc990
\item 272f95fa
\item 6e19193c
\item 44d8ac46
\item 1f642eb9
\item 6e82a1ae
\item ddf7fa4f
\item 543a7ed5
\item f8c80d96
\item 3befdf3e
\item 63613498
\item 98cf29f8
\item a1570a43
\item 83302e8f
\item 0d3d703e
\item a65b410d
\item c0f76784
\item 39e1d7f9
\item d687bc17
\item 7e0986d6
\item 3e980e27
\item ef135b50
\item d2abd087
\item b230c067
\item d43fd935
\item 776ffc46
\item a5313dff
\item 264363fd
\item 6c434453
\item 00d62c1b
\item 06df4c85
\item a79310a0
\item 5c0a986e
\item 444801d8
\item 25ff71a9
\item d06dbe63
\item 6d75e8bb
\item 3345333e
\item 50846271
\end{enumerate}
\end{multicols}

\end{document}